\documentclass[11pt,a4paper]{article}
\usepackage[hyperref]{ranlp2023}
\usepackage{times}
\usepackage{latexsym}

\usepackage{microtype}
\usepackage{multirow}

\usepackage{subcaption}

\aclfinalcopy 
\def\aclpaperid{***} 

\usepackage{comment}
\usepackage{amssymb}
\usepackage{booktabs}
\usepackage{color}
\usepackage{graphicx}
\usepackage{tabularx}
\usepackage{pifont}
\newcommand{\xmark}{\ding{55}}%

\title{\textsc{3D-ex}: A Unified Dataset of Definitions and Dictionary Examples}

\author{Fatemah Almeman$^{*\triangle}$ \qquad Hadi Sheikhi$^{\circ}$ \qquad Luis Espinosa-Anke$^{*\diamondsuit}$ \\
         $^{*}$CardiffNLP, School of Computer Science and Informatics, Cardiff University, UK \\
         $^{\triangle}$ College of Computer and Information Sciences, Princess Nourah bint Abdulrahman University, KSA \\
         $^{\circ}$ School of Computer Engineering, Iran University of Science and Technology, Iran\\
         $^{\diamondsuit}$AMPLYFI, UK\\ 
         \texttt{\{almemanf, espinosa-ankel\}@cardiff.ac.uk}\\
         \texttt{ha\_sheikhi@comp.iust.ac.ir}}

\date{}

\begin{document}
\maketitle
\begin{abstract}
Definitions are a fundamental building block in lexicography, linguistics and computational semantics. In NLP, they have been used for retrofitting word embeddings or augmenting contextual representations in language models. However, lexical resources containing definitions exhibit a wide range of properties, which has implications in the behaviour of models trained and evaluated on them. In this paper, we introduce \textsc{3D-ex}, a dataset that aims to fill this gap by combining well-known English resources into one centralized knowledge repository in the form of $<$term, definition, example$>$ triples. \textsc{3D-ex} is a unified evaluation framework with carefully pre-computed train/validation/test splits to prevent memorization. We report experimental results that suggest that this dataset could be effectively leveraged in downstream NLP tasks. Code and data are available at \url{https://github.com/F-Almeman/3D-EX}.

\end{abstract}

\section{Introduction}

Lexicographic definitions have played an important role in NLP. For example, definitions, and more specifically, term-hypernym pairs occurring in them, constitute a core component in applications such as taxonomy learning \cite{navigli2011graph,velardi2013ontolearn,espinosa2016extasem}, knowledge base construction \cite{bovi2015large}, or for augmenting language models (LMs) \cite{joshi2020contextualized,chen2022dictbert}. For this reason, numerous works have proposed methods to extract definitions from corpora (definition extraction, or DE) \cite{navigli2010learning,anke2018syntactically,spala2020semeval}. However, DE, traditionally framed as a sentence classsification problem, plateaus quickly in terms of its applicability to real-world settings for a number of reasons, namely: (1) it is tied to a reference corpus; (2) it does not handle flexible contexts (e.g., definitional information appearing across several sentences); and (3) incorporating monolithic sentence-level definitional knowledge into LMs during pretraining is not straightforward. A complementary task to the above is definition modeling (DM), a promising direction both from resource creation and NLP standpoints. DM is the task of automatically generating human-readable lexicographic definitions or glosses given some input. From its inception, where \citet{no:17} trained a bidirectional LSTM on $\left<t,d\right>$ pairs, where $t$ is an input term, and $d$ is its corresponding definition, more recent contributions in this area have leveraged contextualized representations by augmenting $t$ with some context $c$ \cite{ni:17,ga:18,is:19,reid2020vcdm,be:20}.

A crucial prerequisite for enabling, among others, successful DM systems is having access to datasets that combine terms, definitions, and \textit{good dictionary examples} \cite{ad:08, ko:19, fr:19}. In lexicographic resources, these good dictionary examples are written by professional lexicographers or domain experts, and often adhere to some style guidelines. This makes these sentences a valuable contextual resource for understanding the meaning of words, sometimes complementing knowledge gaps that may still exist even after reading a concept's definition. 

DM is, arguably, one of the most recent direct NLP application of lexical resources
. We therefore argue for the need of a centralized repository that could be used to train and test DM systems, explore out-of-domain generalization, and most importantly, act as a unified test bed for lexical semantics tasks. In this paper, we fill this gap by introducing \textsc{3D-Ex}, a dataset that unifies a diverse set of English dictionaries and encyclopedias. 
Our results suggest that, indeed, \textsc{3D-Ex} is a valuable resource for testing generative models in lexicographic contexts due to its varied sources, which makes it hard to memorize, and is also helpful for augmenting competitive baselines in downstream tasks.

\section{Related work}
\label{sec:relatedwork}

Lexical resources have a long-standing tradition in lexical semantics \cite{camacho2018interplay}. Given the breadth of the area, we will review some of the most prominent existing resources, and then focus on how these resources have been leveraged in NLP tasks. 

\subsection{Lexical resources}
\label{sec:lrs}

Arguably, the best known lexical resource in NLP is WordNet (WN) \cite{miller1995wordnet}, and as \citet{hovy2013collaboratively} described it, ``the list papers using WN seems endless''. 
Other resources which have complemented or augmented WN in the NLP space include knowledge bases such as Yago \cite{suchanek2008yago}, DBPedia \cite{auer2007dbpedia}, BabelNet \cite{navigli2012babelnet} or WikiData \cite{vrandevcic2014wikidata}\footnote{Note that all these resources include definitions, unlike other resources designed for different purposes such as commonsense reasoning (e.g., ConceptNet \cite{speer2012representing}).}. Traditional dictionaries have also played an important role in NLP, we review these in Section \ref{sec:datasets}, as they constitute the backbone of \textsc{3D-Ex}. 


\subsection{Applications in NLP}
\label{sec:applications}


Lexical resources in general, and dictionaries in particular, have played a critical role in recent years for improving (knowledge-rich and organic) NLP systems. For instance \citet{faruqui2014retrofitting} retrofitted word embeddings using semantic relations; \citet{joshi2020contextualized} and \citet{chen2022dictbert} used definitional information to augment pretrained LMs; and \citet{bovi2015large}, \citet{espinosa2016extasem} and \citet{ijcai2022p615} used definitions for generating knowledge bases. In parallel, a  generative avenue mostly revolving around DM has garnered substantial interest, where earlier works used LSTMs \cite{no:17, ga:18, is:19}, and later contributions shifted to LMs \cite{be:20,huang-etal-2021-definition,august-etal-2022-generating}. These works used DM models for downstream tasks like word sense disambiguation (WSD) \cite{navigli2009word}, word-in-context classification \cite{pilehvar-camacho-collados-2019-wic} or specificity-controlled glossary writing. Other works have explored complementary spaces, e.g.,  exemplification modeling (i.e., generating suitable dictionary examples given a word-definition pair) or full-fledged dictionary writing \cite{ba:21,de2023end,sierra2023spanish}.

\subsection{Datasets}
\label{sec:datasets}
Let us review the datasets we integrate into \textsc{3D-ex} and how they have been applied either in lexicography or downstream NLP tasks.

\paragraph{\noindent \textbf{WordNet}:} WN is an electronic lexical database for English that organises words in groups of synonyms called \textit{synsets} \cite{miller1995wordnet,fe:13}. Each synset is described by its definition, surface forms (lemmas), examples of usage (where available), and the relations between synsets, e.g., hypernymy (is-a), meronymy (is-part) or troponymy (manner-of). WN's primary use in NLP is as a sense inventory \cite{ag:07, zh:22, pu:23}. 
    
\paragraph{\noindent \textbf{CHA}:} CHA \cite{ch:19} is an online dataset of words, definitions and dictionary examples from the Oxford Dictionary. It can be considered as a corpus of ``traditional'' dictionary definitions, and has been leveraged for DM by \citet{be:20} and for benchmarking the quality of WN's examples \cite{al:22}.

\paragraph{\noindent \textbf{Wikipedia}:} Wikipedia is an online encyclopedia that is created by various contributors on the web \cite{ya:16}. In this work we used a dataset that is built by \citet{is:19} from Wikipedia and Wikidata and each entry consists of a phrase, description, and example. This dataset is used to evaluate DM approaches that combine distributional and lexical semantics using continuous latent variables \cite{reid2020vcdm}.

\paragraph{\noindent \textbf{Urban}:} Urban Dictionary is a crowd-sourced dictionary for terms that are not typically captured by traditional dictionaries \cite{st:20}. In this work we used URBAN dataset that was created from Urban dictionary by \citet{reid2020vcdm} as a corpus of uncommon and slang words.

\paragraph{\noindent \textbf{Wiktionary}:} Wiktionary is a freely available web-based dictionary that provides detailed information on lexical entries such as definitions, examples of usage, pronunciation, translations, etc. \cite{ba:22}. It has been used as a resource for WSD \cite{ch:11, ma:13}, especially for retrieving WSD examples which augment labeled data for rare senses \cite{te:21} and for non-English tasks \cite{he:12, se:19}.

\paragraph{\noindent \textbf{Webster's Unabridged}:} Webster's Unabridged is a version of Webster's dictionary \cite{no:90} served by the Project Gutenberg initiative \cite{va:09}. It describes English words by providing definitions and notes (where needed).
     
\paragraph{\noindent \textbf{Hei++}:} Hei++ is a dataset that associates human-made definitions with adjective-noun phrases. Since there is no publicly available dataset to evaluate the quality of definition generation models on free phrases,  Hei++ is built by \citeauthor{be:20} using the test split of the HeiPLAS dataset \cite{ha:15}.

\paragraph{\noindent \textbf{MultiRD}:} The MultiRD dataset was created by \cite{le:19} to evaluate a multi-channel reverse dictionary model that has multiple predictors to predict attributes of target words from given input queries. This dataset uses the English dictionary definition dataset created by \citet{hill2016learning} as the training set and three test sets: a \textit{seen} definition set, an \textit{unseen} definition set, and a description set that includes pairs of words and human-written descriptions. For each entry, it also includes morphemes, lexical names and sememes.

\paragraph{\noindent \textbf{CODWOE}:} The CODWOE (Comparing Dictionaries and Word embeddings) SemEval 2022 shared task \cite{mi:22} aimed to compare two types of semantic descriptions, namely dictionary glosses and word embedding representations. This task was applied to multiple languages, 
and one dataset per language was provided. Each dataset contains a list of examples and, subsequently, each example contains the following key fields: identifier (includes the word), gloss, and embedding-related information.

\paragraph{\noindent \textbf{Sci-definition}:} Sci-definition is a dataset constructed for the task of generating definitions of scientific terms with controllable complexity \cite{august-etal-2022-generating}. The definitions are drawn from MedQuAD \cite{as:19} and Wikipedia Science Glossaries\footnote{\url{https://en.wikipedia.org/wiki/Category:Glossaries_of_science}.}. For each term, 10 journal abstracts are provided from S2ORC \cite{ky:20} to allow models to incorporate related scientific knowledge \cite{fa:19, pe:18}.
     
\section{Building \textsc{3D-EX}: Data Cleaning}
\label{sec:datasets}

A prerequisite for unifying the above resources into \textsc{3D-EX}, is to perform a number of preprocessing steps. This process includes: lower-casing; removing special tokens and any noisy characters such as the \texttt{tab} sign; removing entries where their definitions have more than 10\% of non alphanumeric characters; removing entries that have null values either in words or definitions; removing entries where examples are the same as defined terms, and removing duplicate entries within each dataset or split.

\subsection{Dataset-specific cleaning}

While the above steps are applied to all datasets, each individual resource in \textsc{3D-EX} undergoes a specific preprocessing set of steps:

\paragraph{\textbf{Urban:}} since Urban dictionary is built by end-users who are not trained lexicographers, we found that it has number of noisy definitions (typically, too short, or containing a high proportion of emoticons, exclamation marks, and so forth). To handle them, we built a binary classifier based on RoBERTa-base \cite{liu2019roberta} where 4,000 positive examples are randomly sampled from Wiktionary, CHA and WN, and 2,000 negative examples are randomly sampled from Urban. This classifier, which obtains almost perfect accuracy, is then applied to the entirety of the Urban dataset, leaving \textsc{3D-EX} only with Urban entries that are similar to those in more traditional resources, both in content and, more importantly, in style. Table \ref{tab:urban-clf} lists examples of this filtering process, where we can see Urban-specific properties such as colloquialisms (phrasal verbs, personal pronouns, lack of punctuation marks or high proportion of slang/unknown words).

\begin{table}
\centering
\resizebox{\columnwidth}{!}{%
\begin{tabular}{|p{1.5cm}|p{4cm}|p{4cm}|p{1cm}|}
\hline
\textbf{Term} & \textbf{Definition} & \textbf{Example} & \textbf{F.} \\
\hline
baby bentley & a way to describe a beat up old car you wish was a Bentley & Dave calls his beat-up Neon his baby Bentley & 1 \\
\hline
pang & pangers pingerz pang pangs pangs MDMA ecstasy & Hi Marissa, it's Frank Recard calling. I'll be in the neighborhood later on, and I was wondering if maybe you wanted to get some pang pangs & 1 \\
\hline
suckafish & the correct term for one who you think is a sucker, loser, or anything else & Wow, that guy is being a total suckafish & 1 \\
\hline
farblegarb & a lot of random garbage & The signal was disrupted, producing a lot of farblegarb & 0 \\
\hline
citrixify & the process of modifying or altering a computer application for the purpose of publishing the application using Citrix Presentation Server & In order to properly publish that Java-based application, I had to citrixify it so it would run in a seamless window & 0 \\
\hline
axcellent & when something rocks and is excellent & Dude, that new haircut is axcellent & 0 \\
\hline
\end{tabular}%
}
\caption{Examples of Urban entries that were removed vs. retained (labels 1 vs. 0 in column \textbf{F.}).}
\label{tab:urban-clf}
\end{table}

\paragraph{\textbf{Wiktionary:}} Since some definitions in Wiktionary include the time where words were coined (e.g., ``first attested in the late 16th century'' or ``from 16 c''), we deleted them using regular expressions.

\paragraph{\textbf{MultiRD:}} we removed (again, using regular expressions) uninformative definitions such as "see synonyms at" and "often used in the plural".

\paragraph{\textbf{Sci-definition:}} in order to construct the \textbf{Sci-definition} dataset as \texttt{$<$term, definition, example$>$} triples, we took the following steps: from each abstract, we extracted sentences that include the target term, which would act as examples. From these examples, we excluded sentences only containing lists of keywords (typically found in abstracts), and also any example with more than 10\% non alphanumeric characters (similarly to our approach to cleaning definitions in Section \ref{sec:datasets}).

\subsection{Unification and splitting}

\begin{table*}[!t]
\centering
\scalebox{0.8}{
\small
\resizebox{\textwidth}{!}{%
\begin{tabular}{@{}lrrrrr@{}}
\toprule
                              & \multicolumn{1}{c}{\textbf{orig. \#entries}} & \multicolumn{1}{c}{\textbf{cl. \#terms}} & \multicolumn{1}{c}{\textbf{cl. \# \textless{}T,D\textgreater{}}} & \multicolumn{1}{c}{\textbf{cl. \#\textless{}T,D,E\textgreater{}}} \\ \midrule
\textbf{WordNet}      & 44,351   &  20,435  &  36,095 & 44,241  \\                                                             
\textbf{CHA}         & 785,551   & 30,841   & 75,887  & 752,923  \\

\textbf{Wikipedia}   & 988,690   & 162,809  & 167,569  & 960,097  \\

\textbf{Urban}       & 507,638   & 119,016  & 145,574  & 145,896  \\

\textbf{Wiktionary}  & 145,827   & 76,453   & 85,905 & 140,190   \\

\textbf{CODWOE}      & 63,596    & 25,861   & 45,065 & 63,137   \\

\textbf{Sci-definition}  & 8,263 & 5,281  & 6,251  & 166,660 \\ 
\midrule

\textbf{Webster's Unabridged} & \multicolumn{1}{r}{159,123} 
& \multicolumn{1}{r}{89,234}                
& \multicolumn{1}{r}{143,782}                                         & \multicolumn{1}{r}{-}                                               \\
\textbf{MultiRD} & \multicolumn{1}{r}{901,200}          
& \multicolumn{1}{r}{50,460}                
& \multicolumn{1}{r}{671,505}                                        & \multicolumn{1}{r}{-}                                               \\

\textbf{Hei++}                
& \multicolumn{1}{r}{713}                      
& \multicolumn{1}{r}{713}                  
& \multicolumn{1}{r}{713}                                            & \multicolumn{1}{r}{-}                                               \\ \midrule
\textbf{3D-EX}                
& \multicolumn{1}{r}{}                      
& \multicolumn{1}{r}{438,956}                  
& \multicolumn{1}{r}{1,327,342}                                      & \multicolumn{1}{r}{2,268,225}  
\\ \bottomrule
\end{tabular}%
}
}
\caption{Dataset statistics before (orig.) and after (cl.) preprocessing, and in terms of unique entries involving terms (\textbf{T}), definitions (\textbf{D}), examples (\textbf{E}). Aggregated statistics are provided between two sets, datasets with examples (top) and without (bottom). The last row is related to 3D-EX dataset.}
\label{tab:dataset-stat}
\end{table*}

\begin{table}[!th]
\resizebox{\columnwidth}{!}{%
\begin{tabular}{@{}lrrr|rrr|lrrr@{}}
\toprule
               & \multicolumn{3}{c|}{\textbf{Term length}}                                     & \multicolumn{3}{c|}{\textbf{Definition length}} & \multicolumn{3}{c}{\textbf{Example length}} \\ \midrule
               
               & \multicolumn{1}{c}{min.} & \multicolumn{1}{c}{max.} & \multicolumn{1}{c|}{avg.} & \multicolumn{1}{c}{min.} & \multicolumn{1}{c}{max.} & \multicolumn{1}{c|}{avg.} & \multicolumn{1}{c}{min.} & \multicolumn{1}{c}{max.} & \multicolumn{1}{c}{avg.} \\ \midrule
               
WordNet        & 1   & 1   & 1
               & 1   & 52  & 7.50
               & 1  & 46   & 5.77 \\
               
CHA            & 1   & 1   & 1
               & 1   & 71  & 10.31
               & 2   & 141   & 17.86 \\

Wikipedia      & 1   & 16   & 1.84
               & 1   & 32   & 6.012
               & 2   & 40   & 18.70 \\

Urban          & 1   & 31   & 1.47
               & 1   & 32   & 10.01
               & 2   & 42   & 11.45 \\

Wiktionary     & 1   & 10   & 1.22
               & 1   & 100   & 9.24
               & 2   & 288   & 26.52 \\

CODWOE         & 1   & 1   & 1
               & 1   & 114   & 10.86
               & 1   & 214   & 22.26 \\

Sci-definition      & 1   & 11   & 1.70
                    & 2   & 94   & 18.49
                    & 1   & 726   & 25.72 \\

Webster's Unabridged       & 1   & 3   & 1.00
               & 1   & 90   & 9.19
               & -  &  -  & - \\

MultiRD        & 1   & 1  & 1
               & 1   & 144   & 11.72
               & -    &  -   &  -  \\

Hei++          & 2   & 2   & 2
               & 3   & 23   & 8.12
               & -  &  -  &  - \\

\bottomrule
\end{tabular}%
}
\caption{Length statistics per dataset after cleaning.}
\label{tab:len_stat}
\end{table}

\begin{table*}[!t]
\small
\resizebox{\textwidth}{!}{%
\begin{tabular}{p{3cm} p{4cm} p{5cm} p{2cm}}
\toprule
\multicolumn{1}{c}{\textbf{Term}} & \multicolumn{1}{c}{\textbf{Definition}}                                                                                                                                                   & \multicolumn{1}{c}{\textbf{Example}}                                                                                                                                                                    & \multicolumn{1}{c}{\textbf{source}} \\ \midrule
emergent                          & coming into existence                                                                                                                                                                     & an emergent republic                                                                                                                                                                                    & WordNet                             \\ \midrule
word                              & an (order; a request or instruction); an expression of will                                                                                                                               & he sent word that we should strike camp before winter                                                                                                                                                   & Wiktionary                          \\\midrule
central london                    & innermost part of london , england                                                                                                                                                        & westminster is an area of central london within the city of westminster , part of the west end , on the north bank of the river thames                                                                  & Wikipedia                           \\\midrule
ejac-flashback                    & when a picture or video is familiar to you                                                                                                                                                & dude I've just had a ejac-flashback that chick was last nights wank material                                                                                                                            & Urban                               \\\midrule
notice                            & a displayed sheet or placard giving news or information                                                                                                                                   & look out for the notice of the samaritans information evening in the end of september                                                                                                                   & CHA                                 \\\midrule
worship                           & to participate in religious ceremonies                                                                                                                                                    & we worship at the church down the road                                                                                                                                                                  & CODWOE                              \\\midrule
accessory navicular bone          & an accessory navicular bone is a small bone located in the middle of the foot                                                                                                             & the accessory navicular bone is one of the most common accessory ossicles, which sometimes become symptomatic                                                                                           & Sci-definition                      \\\midrule
able                             & having sufficient power, strength, force, skill, means, or resources of any kind to accomplish the object                                                                                                                    & -                                                                                                                                                                                                     & Webster's Unabridged                \\\midrule
abbreviation                      & an abbreviation is a shorter way to write a word or phrase                                                                                                                                & -                                                                                                                                                                                                     & MultiRD                             \\\midrule
skew picture                      & an inaccurate or partial representation of a situation                                                                                                                                    & -                                                                                                                                                                                                     & Hei++                               \\ \bottomrule
\end{tabular}%
}
\caption{Examples of entries available in 3D-EX. }
\label{tab:examples}
\end{table*}

Tables \ref{tab:dataset-stat} and \ref{tab:len_stat} show summary statistics for each dataset. It is desirable to keep a reference to the original source (dictionary or glossary) for each entry, however, we noticed that there are \texttt{$<$term, definition, example$>$} duplicates across datasets. This is why the final \textsc{3D-ex} resource contains the \textsc{source} field as an array containing the sources where that entry was found. Furthermore, in terms of splitting \textsc{3D-ex} for experimentation, it is well known that an issue in word/phrase classification datasets can occur due to a phenomenon known as ``lexical memorization'' \cite{levy2015supervised}, where supervised models tend to associate prototypical features to word types. This has been typically been addressed by releasing two splits, one random, and one known as ``the lexical split'', where all instances of a given term do not appear across splits \cite{vulic2017hyperlex,apidianaki2021all,anke2022multilingual}. We follow this practice and release \textsc{3D-ex} with a Random and a Lexical split. Tables \ref{tab:examples} and \ref{tab:split_stat} show examples of entries in \textsc{3D-ex} and dataset statistics after unification in terms of unique instances across both splits, respectively.

\begin{figure}[h]
  \centering
  \begin{subfigure}{\linewidth}
    \centering
    \includegraphics[width=\linewidth]{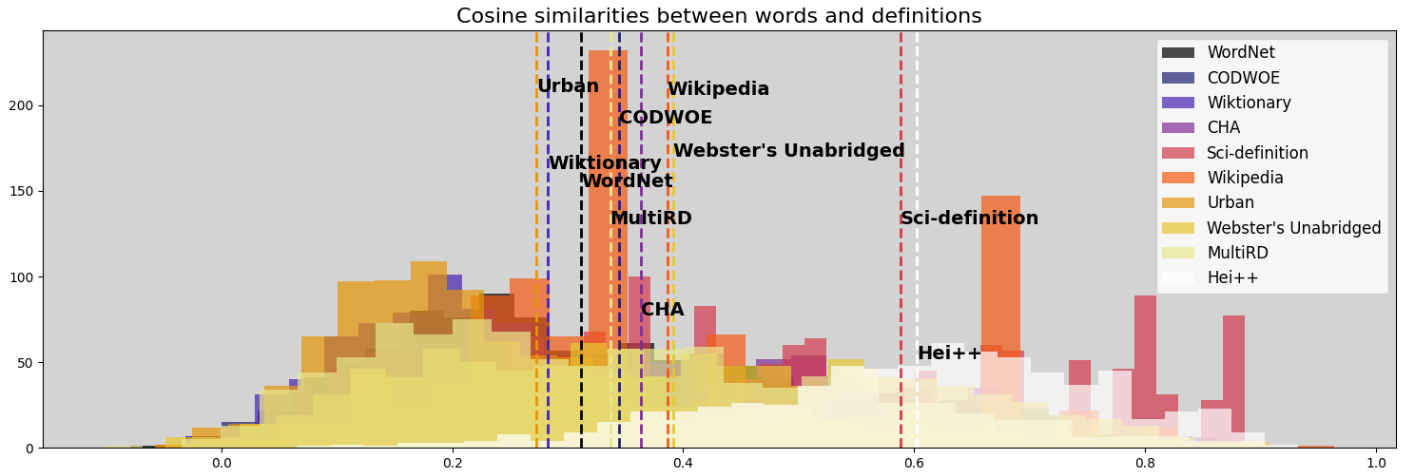}
    \caption{Word-definition comparison}
    \label{fig:histwd}
  \end{subfigure}

  \begin{subfigure}{\linewidth}
    \centering
    \includegraphics[width=\linewidth]{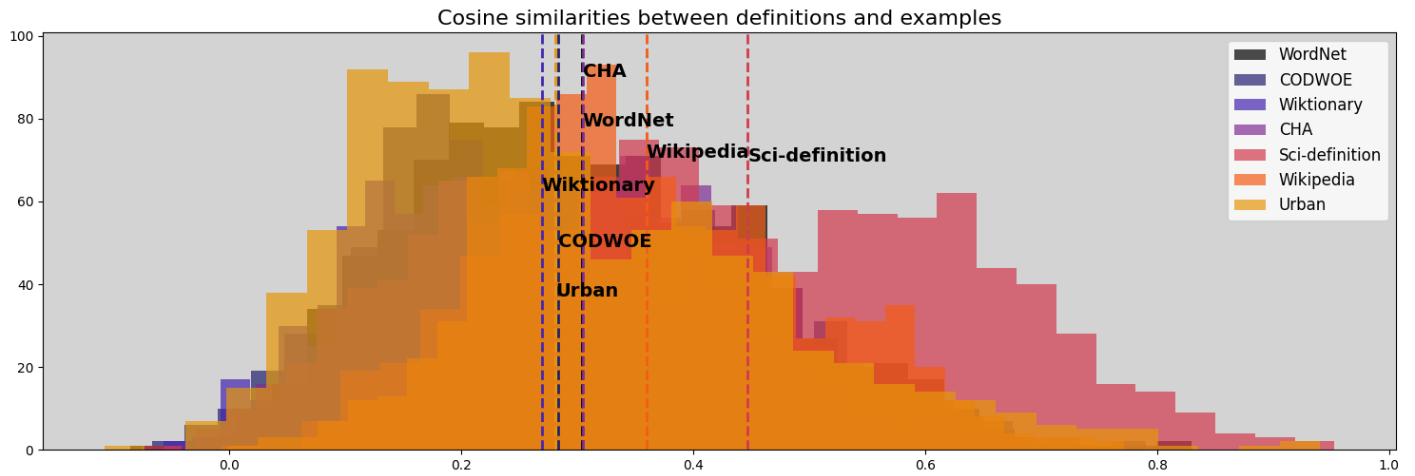}
    \caption{Definition-example comparison}
    \label{fig:histde}
  \end{subfigure}
  \caption{Histograms with SBERT-based cosine similarities of the datasets in \textsc{3D-ex}.}
  \label{fig:hist}
\end{figure}

Finally, to shed some light on how similarities are distributed across datasets, we investigate cosine similarities of their SBERT embeddings, and compute similarities between terms and definitions, and between definitions and terms (see Figure \ref{fig:hist}). An immediate finding by inspecting these similarities is that Hei++, a carefully curated dataset used to evaluate multiword DM systems, is the one showing the highest similarity between terms and definitions (Figure \ref{fig:histwd}), this is likely because, first, entries in Hei++ are rather specific, and do not include generic and frequently used terms. This, along with, also, a rather detailed definition, makes their similarity rather high. On the opposite end of the spectrum we unsurprisingly find Urban dictionary, although it remains for future work to explore whether Urban Dictionary's definitions are indeed dissimilar to their corresponding terms, or because they are so rare that their embeddings are of lower quality. Interestingly, we also find that Sci-definition also exhibits high similarity between terms and definitions. Concerning definitions and examples (Figure \ref{fig:histde}), Sci-definition is again the one with the highest similarity scores, and interestingly, Wiktionary is the dictionary with the lowest aggregate similarity, which suggests that examples in Wiktionary could be purposefully written to cover different topics than their definitions. As with the case of Urban Dictionary, a careful semantic analysis of these dictionaries remains for future work.

\begin{table}[!t]
\LARGE
\renewcommand{\arraystretch}{1.2}
\resizebox{\columnwidth}{!}{%
\begin{tabular}{@{}lrrr|rrr@{}}
\toprule
                 & \multicolumn{3}{c|}{\textbf{Random split}}                                                                        & \multicolumn{3}{c}{\textbf{Lexical split}}                                                                       \\ \midrule
                 & \multicolumn{1}{c}{\textbf{train}} & \multicolumn{1}{c}{\textbf{validation}} & \multicolumn{1}{c|}{\textbf{test}} & \multicolumn{1}{c}{\textbf{train}} & \multicolumn{1}{c}{\textbf{validation}} & \multicolumn{1}{c}{\textbf{test}} \\
WordNet     & 26,603  &  8,788   & 8,850 &  27,053 & 8,573 & 8,793 \\

CHA      & 451,191 &15,1338 & 50,394 & 452,321 & 157,847 & 143,949  \\

Wiktionary   & 84,111 & 28,127 & 27,952 & 89,607 & 29,176 & 23,832   \\

Wikipedia    & 575,554 & 197,697 & 186,846 & 505,964 & 240,781 & 213,379 \\

Urban      & 87,429 & 29,142 & 29,325 & 91,239 & 29,783 & 24,881 \\

CODWOE      & 37,774 & 12,755 & 12,608 & 39,737 & 12,609 & 13,166  \\

Sci-definition  & 101,129 & 31,766 & 33,765 & 106,175 & 35,966 & 24,519\\
 \midrule

Webster's Unabridged   & 84,802 & 28,213 & 28,221 & 93,423 & 30,198 & 19,696 \\

MultiRD     & 384,295 & 127,580 & 128,178 & 404,114 & 125,072 & 112,948 \\

Hei++       & 426 & 152 & 135 & 428 & 143 & 142       

\\ \bottomrule
\end{tabular}%
}
\caption{Breakdown of \textsc{3D-ex} unique entries per split type (random and lexical) and per split. Note that unique entries consist of $<$term,def.,example,source$>$ (first 6 rows) or $<$term,def.,source$>$ (bottom 3 rows).}
\label{tab:split_stat}
\end{table}

\begin{table}[!th]
\resizebox{\columnwidth}{!}{%
\begin{tabular}{@{}lrrr|rrr@{}}
\toprule
               & \multicolumn{3}{c|}{\textbf{Random Split}}                                     & \multicolumn{3}{c}{\textbf{Lexical Split}}                                    \\ \midrule
               & \multicolumn{1}{c}{prec.} & \multicolumn{1}{c}{rec.} & \multicolumn{1}{c|}{f1} & \multicolumn{1}{c}{prec.} & \multicolumn{1}{c}{rec.} & \multicolumn{1}{c}{f1} \\ \midrule
WordNet        & 0.73 & 0.23 & 0.35 & 0.33 & 0.05 & 0.09  \\
CHA            & 0.65 & 0.48 & 0.55 & 0.64 & 0.47 & 0.54  \\
Wiktionary     & 0.80 & 0.53 & 0.64 & 0.65 & 0.33 & 0.44  \\
Wikipedia      & 0.98 & 0.97 & 0.98 & 0.97 & 0.97 & 0.97  \\
Urban          & 0.94 & 0.87 & 0.91 & 0.97 & 0.66 & 0.79  \\
CODWOE         & 0.93 & 0.55 & 0.69 & 0.92 & 0.42 & 0.58  \\
Sci-definition & 0.99 & 0.99 & 0.99 & 0.99 & 0.99 & 0.99  \\
Webster's Unabridged        & 0.82 & 0.70 & 0.76 & 0.75 & 0.63 & 0.68   \\
MultiRD        & 0.89 & 0.90 & 0.89 & 0.84 & 0.91 & 0.88   \\ 
Hei++          & 0    &   0  & 0    & 0    &  0   &  0     \\ \midrule
Average        & 0.77 & 0.62 & 0.68 & 0.71 & 0.54 & 0.60  \\ \bottomrule
\end{tabular}%
}
\caption{Results in the source classification experiment, reported both for the Random and Lexical splits of \textsc{3D-EX}.}
\label{tab:source-clf}
\end{table}

\section{Experiments and Results}

In order to test the usefulness of \textsc{3D-ex}, we perform an intrinsic set of experiments where we ``stress test'' the dataset for artifacts, indirect data leakage (near-synonyms), potential for memorization, etc. This, we argue, is an important step to guarantee \textsc{3D-ex} can be used for testing lexical semantics models based on it.

\subsection{Source classification}

In the task of \textit{source classification}, the goal is to, given a $<$term,definition$>$ instance, predict its original source. We posit that this is an important experiment to determine which sources are more unique (i.e., easier to classify), and which seem to conflate different lexicographic features (e.g., writing style, coverage or any other artifact). To this end, we fine-tune \texttt{roberta-base} \cite{liu2019roberta} for 3 epochs on the training set of \textsc{3D-ex}. Note that this is a 9-way multilabel classification problem, since for a given $<$term,definition$>$ tuple, there may be more than one associated source.

We report the results of this experiment in Table \ref{tab:source-clf}. We can see how the lexical split is substantially harder than the random split.


\subsection{Reverse dictionary}

Reverse dictionary (or concept finder) is a helpful application for copywriters, novelists, translators seeking to find words or ideas that might be ``on the tip of their tongue'' \cite{hill2016learning}. It is also reflection of the interactions between a speaker and the mental lexicon \cite{zock2004word,zock2010deliberate}. More relevant to NLP, however, reverse dictionary datasets can be seen as benchmarks for evaluating representation learning methods, as there are works that have used definitions as, e.g., the sole source for learning word embeddings \cite{bosc2017learning} or for debiasing them \cite{kaneko2021dictionary}. 

This task is a ranking problem in which, given a definition, the task is to retrieve a ranked list of the most relevant words, and it has a long-standing tradition in computational semantics \cite{bi:04, do:02, el:04, gl:92, th:16} . To establish a set of baseline results on this task, we report results from several embedding models on the random and lexical test sets. Note that while these baselines are unsupervised, we only report results on the test sets to accommodate future experiments by supervised systems. In terms of evaluation, we report \textit{Mean Reciprocal Rank} (MRR), which rewards the position of the first correct result in a ranked list of outcomes:

$$
\mbox{{MRR}} = \frac{1}{|Q|}\sum_{i=1}^{|Q|}\frac{1}{rank_i}
$$

\noindent where $Q$ is a sample of experiment runs and $rank_i$ refers to the rank position of the \textit{first} relevant outcome for the \textit{i}th run. MRR is commonly used in Information Retrieval and Question Answering, but has also shown to be well suited for lexical semantics tasks such as collocation discovery \cite{Wu-etal10,rodriguez2016semantics}.

We evaluate the performance of traditional sentence encoding SBERT \cite{reimers2019sentence} models, namely {\small{\texttt{all-MiniLM-L6-v2}}
}, {\small{\texttt{all-distilroberta-v1}}} and {\small{\texttt{all-mpnet-base-v2}}}. We also evaluate Instructor \cite{su2022one}, an instruction-based encoder that can generate text embeddings tailored to any task given the appropriate prompt. Instructor works by optionally providing the type of the target text (e.g., ``a Wikipedia sentence'') and the task (e.g., ``document retrieval''), to ultimately build a prompt such as ``Represent this Wikipedia sentence for retrieving relevant documents''. For our use case, we test three variants of Instructor for encoding both words and definitions: (1) no instruction; (2) providing a generic description of the target text (i.e., ``the sentence'' and ``the word''); and (3) providing a domain-specific description of the target texts (i.e., ``the dictionary definition'' and ``the dictionary entry'').

We show the results of the SBERT models in Table \ref{tab:sbert}, and the Instructor results in Table \ref{tab:instructor}. We can see that even without any instruction prepended to the embedder, the Instructor model outperforms vanilla SBERT models, and that, interestingly, the best results overall in both splits (random and lexical) are obtained by providing a generic description of target words, and in the random split it is better to not include instructions for the definitions, while in the lexical split the best performing configuration involves providing detailed instructions for embedding the \textsc{3D-ex} definitions.

As a final piece of analysis, we perform experiments on both test sets with the best performing model (based on the split type) to see which sources are harder to solve in the task of reverse dictionary. From Table \ref{tab:reverse-dict-datasets}, it can be seen that Wikipedia and Urban are the most challenging resources for this task, which could be attributed to either or both dataset size and large number of very similar definitions and terms, as opposed to for instance Hei++ or Sci-definition, which are meant to capture unique terms. These are, by nature, more unique when compared to the rest of the lexicon, an insight we revealed when exploring dataset-specifc similarities in Figure \ref{fig:hist}.

\begin{table}[]
\tiny
\resizebox{\columnwidth}{!}{%
\begin{tabular}{@{}lrr@{}}
\toprule
\multicolumn{1}{c}{Model} & \multicolumn{1}{c}{Random} & \multicolumn{1}{c}{Lexical} \\ \midrule
all-distilroberta-v1      & 8.41      &    11.38          \\
all-MiniLM-L6-v2          & 9.40      &    13.75           \\
all-mpnet-base-v2         & 10.98     &    15.34           \\ \bottomrule
\end{tabular}%
}
\caption{Reverse Dictionary results of the SBERT models on the reverse dictionary task in the two \textsc{3D-ex} test sets.}
\label{tab:sbert} 
\end{table}

\begin{table}[]
\begin{tabular}{ll|rrr}
\toprule
\multicolumn{2}{c|}{\multirow{2}{*}{Random}}  & \multicolumn{3}{c}{word}                                                      \\ \cline{3-5} 
\multicolumn{2}{c|}{}                         & \multicolumn{1}{c}{no} & \multicolumn{1}{c}{gen.} & \multicolumn{1}{c}{dict.} \\ \hline
\multirow{3}{*}{definition} & no    & 14.18 & \bf{14.71} & 14.56\\
                            & gen.  & 13.64 & 14.07 & 14.06 \\
                            & dict. & 14.19 & 14.59 & 14.57  \\ \midrule
\multicolumn{2}{c|}{\multirow{2}{*}{Lexical}} & \multicolumn{3}{c}{word}                                                      \\ \cline{3-5} 
\multicolumn{2}{c|}{}                         & \multicolumn{1}{c}{no} & \multicolumn{1}{c}{gen.} & \multicolumn{1}{c}{dict.} \\ \hline
\multirow{3}{*}{definition}  & no    & 19.16 & 20.25 & 20.02 \\
                             & gen.  & 18.70 & 20.04 & 19.86 \\
                             & dict. & 19.64 & \bf{20.82} & 20.60 \\ \bottomrule
\end{tabular}
\caption{MRR Results on Reverse Dictionary leveraging Instructor Embeddings when using no instruction (no), generic (gen.) or tailored to the task (dict.).}
\label{tab:instructor}
\end{table}

\begin{table}[!h]
\centering
\begin{tabular}{@{}lrr@{}}
\toprule
\multicolumn{1}{c}{Dataset} & \multicolumn{1}{c}{Random} & \multicolumn{1}{c}{Lexical} \\ \midrule
WordNet                     &  32.97 &	42.27              \\
Wiktionary                  &  50.65 &  53.05                \\
Wikipedia                   &  9.25  &  9.19                   \\
Urban                       &  18.47 &  17.49                 \\
CODWOE                      &  39.74 &  46.89                \\
CHA                         &  30.82 &  35.86                  \\
Sci-definition              &  82.38 &	82.53                 \\
Webster's Unabridged        &  30.53 &	34.11                  \\
MultiRD                     &  16.69 &	27.41                  \\
Hei++                       &  96.79 &	94.49                  \\ \bottomrule
\end{tabular}%
\caption{Breakdown of the reverse dictionary results in terms of MRR for the two test sets (random and lexical) in 3D-EX.}
\label{tab:reverse-dict-datasets}
\end{table}

\section{Conclusions and future work}

In this paper we have introduced \textsc{3D-EX}, a dataset that unifies different encyclopedias and dictionaries into one single resource. We have conducted an in-depth analysis of the dataset across several splits (random vs lexical), as well as dictionary source classification and reverse dictionary experiments. Our results suggest that this dataset is both challenging for representation learning methods and promising as a resource for augmenting lexical semantics systems. It has also helped us unveil semantic properties in the different dictionaries and encyclopedias we have integrated into \textsc{3D-EX}. 

For the future, we would like to further explore the potential of \textsc{3D-EX} for downstream NLP tasks, incorporating more resources, and exploring multilingual variants. An additional avenue would be to explore the interaction of unorthodox dictionaries like Urban with traditional lexicographic resources in the context of controlled technical/jargon DM. Finally, leveraging \textsc{3D-EX} as a resource for pretraining LMs, similarly to the DictBERT approach \cite{chen2022dictbert}, could help inform LMs with new, domain-specific and/or colloquial terms.

\section*{Ethics and Broader Impact Statement}
This paper is concerned with the automatic building of a dataset by combining publicly available information in the web. As a result, there could be potential for the presence of incorrect or harmful information in this derived dataset, especially if crowdsourced; however, we encourage collaborative efforts from the community to help address these risks. Specifically, vulgar, colloquial, or potentially harmful information in Urban Dictionary, which the authors of this paper do not endorse. 

\bibliographystyle{acl_natbib}
\bibliography{ranlp2023}


\end{document}